\documentclass[10pt,twocolumn,A4paper]{article} 

\usepackage{iccv}
\usepackage{times}
\usepackage{epsfig}
\usepackage{graphicx}
\usepackage{amsmath}
\usepackage{amssymb}
\usepackage{xcolor}

\usepackage[pagebackref=false,breaklinks=true,letterpaper=true,colorlinks,bookmarks=false]{hyperref}
\usepackage{authblk}

\usepackage{adjustbox}
\usepackage{subcaption}    

\usepackage{multirow}
\usepackage{hhline}

\usepackage{color, colortbl}
\definecolor{Gray}{gray}{0.9}

\iccvfinalcopy

\begin{document}

\title{Guided Super-Resolution as Pixel-to-Pixel Transformation}
\date{}

\author{Riccardo de Lutio}
\author{Stefano D'Aronco}
\author{Jan Dirk Wegner}
\author{Konrad Schindler} 
\affil{EcoVision Lab, Photogrammetry and Remote Sensing, ETH Z\"urich}

          


%


    \maketitle

\begin{abstract}
   Guided super-resolution is a unifying framework for several computer vision tasks where the inputs are a low-resolution \emph{source image} of some target quantity (e.g., perspective depth acquired with a time-of-flight camera) and a high-resolution \emph{guide image} from a different domain (e.g., a grey-scale image from a conventional camera); and the target output is a high-resolution version of the source (in our example, a high-res depth map). The standard way of looking at this problem is to formulate it as a super-resolution task, i.e., the source image is upsampled to the target resolution, while transferring the missing high-frequency details from the guide. Here, we propose to turn that interpretation on its head and instead see it as a pixel-to-pixel mapping of the guide image to the domain of the source image. The pixel-wise mapping is parametrised as a multi-layer perceptron, whose weights  are learned by minimising the discrepancies between the source image and the downsampled target image. Importantly, our formulation makes it possible to regularise only the mapping function, while avoiding regularisation of the outputs; thus producing crisp, natural-looking images. The proposed method is unsupervised, using only the specific source and guide images to fit the mapping. We evaluate our method on two different tasks, super-resolution of depth maps and of tree height maps. In both cases we clearly outperform recent baselines in quantitative comparisons, while delivering visually much sharper outputs.
\end{abstract}


\section{Introduction}

\begin{figure}
    \centering
    \includegraphics[trim={0 0 7cm 0},clip,scale=0.48]{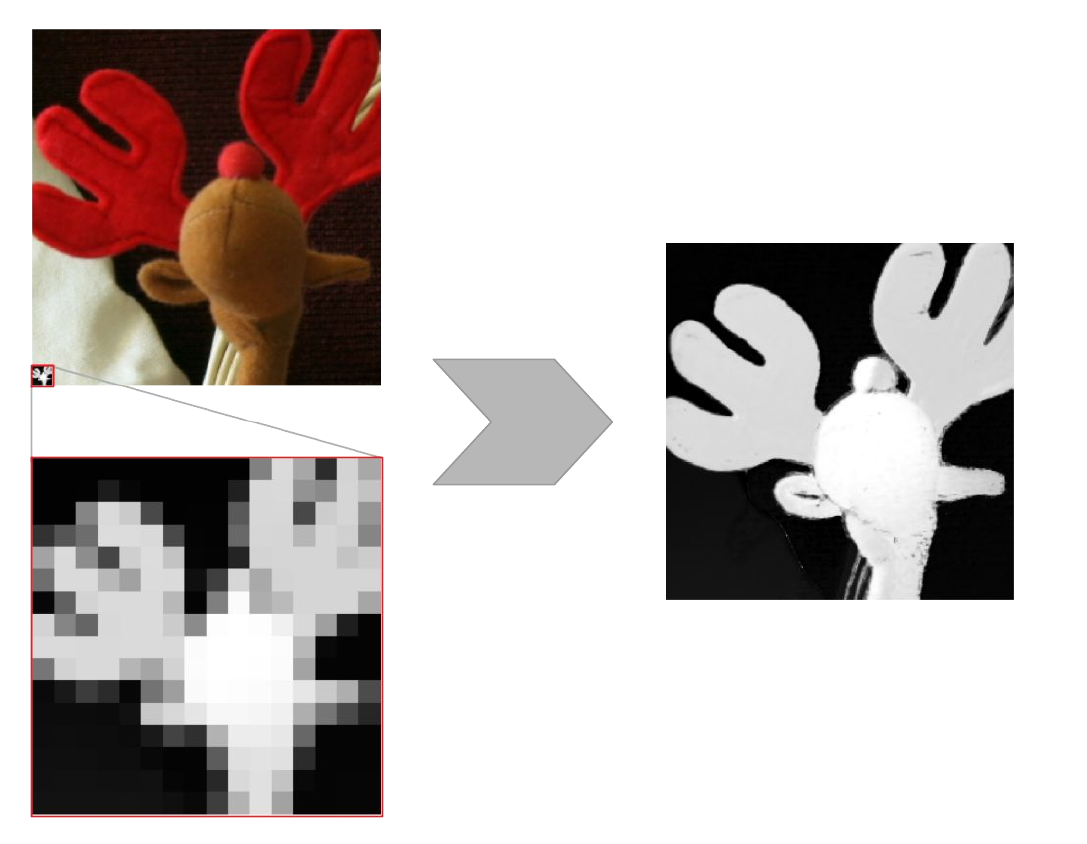}
    \caption{Guided super-resolution: given a low-resolution depth map and a high-resolution guide image, our method predicts a high-resolution depth map. The figure shows an example output of the proposed method, for an upsampling factor of $16 \times$.}
    \label{fig:teaser}
\end{figure}

A number of computer vision tasks can be seen as instances of \emph{guided super-resolution}. For instance, many robots are equipped with a conventional camera as well as a time-of-flight camera (or a laser scanner). The latter acquires depth maps of low spatial resolution, respectively large pixel footprint in object space, and it is a natural question whether one can enhance its resolution by transferring details from the camera image -- see Figure~\ref{fig:teaser}. Another example is environmental mapping, where maps of parameters like tree height or biomass are available at a mapping resolution that is significantly lower than the ground sampling distance of modern earth observation satellites. 

An alternative view of guided super-resolution is as a generalisation of guided filtering~\cite{He2013}, widely used in image processing and analysis. A guided filter maps a source image to a target image of \emph{the same size} by computing, at each pixel, a function that depends on the local neighbourhood in both the source and the guide image (which can be the source itself, as in the popular bilateral filter \cite{Tomasi1998}).
Guided super-resolution does the same, except that the source image has a lower spatial resolution and must additionally be upsampled in the process.

The standard way to model guided super-resolution is as an inverse problem: the source image is understood as the result of downsampling the target image. The objective is to undo that operation, utilising the guide to constrain the solution, by transferring high-frequency details that were lost during downsampling, such as fine structures and sharp boundaries.
Model inference can either be done by directly minimising an appropriate loss function, e.g., with variational methods~\cite{Ferstl2013}; or in two separate steps, e.g., upsampling with generic bilinear or bicubic interpolation followed by guided filtering~\cite{Yang2007}. 

Here, we propose an alternative interpretation of guided super-resolution, where the roles of the source and guide images are swapped: rather than finding a transformation from source to target and constraining the output to be consistent with the guide from a different image domain; we instead prefer to find a transformation from the guide to the target, i.e., a pixel-wise mapping from one image domain to another \emph{without changing the resolution}, and constrain the output by demanding that its downsampled version matches the source image.

In our implementation, we parametrise the mapping as a multi-layer perceptron that takes as input all channels of the guide image \emph{at a single pixel} (plus two additional "channels" corresponding to the pixel's $x$- and $y$-coordinates). In CNN terminology, the guide is augmented with two extra channels that encode pixel location, and then passed through a convolutional network whose layers all use only $1\times 1$ kernels. Thus, the transformation from the guide to the target domain acts on pixels individually, without looking at their neighbours. Spatial context relations are encoded implicitly, and adaptively per image, via the structural bottleneck created by learning a single set of transformation parameters that must be valid for all pixels. We refer to this setup as "pixel-to-pixel transformation", as opposed to "image-to-image translation" with large receptive fields. 
Importantly, our method is \emph{unsupervised}: while the mapping is structurally a form of CNN, we do not learn a static set of network weights from a training set and then apply those weights to every new test image. Rather, we fit an individual set of weights for each new image similarly to \cite{Ulyanov2018}, using all its pixels as "training data" and the consistency with the low-resolution source as "supervision". 

We argue that this view of guided super-resolution has two very practical advantages. 
\emph{(i)} by starting already at the desired resolution, and using only $1\times 1$ kernels, \emph{different input locations do not mix}, which avoids blurring.
\emph{(ii)} by using the same mapping function for all pixels and placing a shrinkage prior on its parameters, one obtains a well-posed problem \emph{without regularisation of the output image}. In this way,  blurring is also avoided at the output stage.
Together, these properties lead to outputs with superior sharpness.

The \textbf{contribution} of this paper is a novel formulation of guided super-resolution, as unsupervised learning of a pixel-to-pixel transformation from the guide to the target image, constrained by the low-resolution source. We present experiments on two tasks: super-resolution of depth maps, and super-resolution of tree height maps. They demonstrate that our formulation clearly outperforms competing super-resolution methods at high upsampling factors ($\times8$ to $\times32$).

\section{Related Work}

\paragraph{Guided filtering.} A large body of work exists about guided filtering, without the additional challenge of super-resolution. The general principle is to enhance the source image by applying a filter whose output depends not only on a local neighbourhood of the source image, but also on weights derived from the same neighbourhood in the guide image. The starting point is the  bilateral filter \cite{Tomasi1998}, where the source image itself serves as a guide. Classical examples that employ a guide from a different domain include the joint bilateral filter \cite{Petschnigg2004},  the guided filter (GF) \cite{He2013} and the weighted median filter \cite{Ma2013}.
Guided filtering has been used to a diverse range of image processing applications, ranging from low-level tasks like denoising \cite{He2013} or colourisation \cite{Kopf2007} all the way to stereo matching \cite{hosni2013}.

\paragraph{Guided super-resolution.} Extensions of guided filtering to the super-resolution problem have been explored for super-resolving depth, as well as for low-level operations like tone mapping and image colourisation. We distinguish between \emph{local methods} based on the above local filtering principle, and \emph{global methods} that formulate the upsampling task as a global energy minimisation.

The local methods are variants of the two-step procedure, i.e., first upsample the low-resolution source image with naive interpolation, then enhance it by applying a filter that is controlled by the high-resolution guide \cite{Kopf2007, Yang2007}. Variants include using the geodesic distance in the high-res image instead of the raw contrast \cite{Liu2013}, and combining the contrast in both the source \emph{and} the guide image to determine the filter strength \cite{Chan2008}.

Global methods formulate the super-resolution as an energy minimisation problem, whose solution returns the values of all pixels in the target image. The energy function consists of a data term that measures the compatibility between the downsampled target image and the low-resolution source image, and a smoothness term to regularise the ill-posed problem. In the guided scenario the latter term is not an isotropic preference for smooth solutions, but is modulated by the guide image.
The global approach has been implemented as a Markov Random Field \cite{Diebel2006}, and has been extended to additionally include the idea of non-local means to enhance image structures \cite{Park2011}. Another
possible implementation is as variational inference \cite{Ferstl2013}, with an anisotropic version of the total generalised variation (TGV) prior, modulated by the guide image. It has also been proposed to replace the TGV prior by an auto-regressive model \cite{Yang2014}, whose parameters are again a function of the bilateral filter response in the guide image.

Recently some methods have appeared that embed the idea of bilateral/guided filtering in a global optimisation framework, rather than apply a local filter. In particular, the fast bilateral solver (FBS) \cite{Barron2016} offers an optimisation mechanism based on a sparse linear system \cite{Barron2015} to obtain bilateral-smooth outputs with sharp discontinuities. Whereas the static/dynamic (SD) filter \cite{Ham2018} converts the guided filtering problem into a non-convex optimisation  that is solved by the majorisation-minimisation algorithm. Both have been successfully used for guided super-resolution, besides other image processing tasks.

\paragraph{Learned guided super-resolution.} 
The methods described so far are unsupervised. There is also a line of work that learns from examples how to upsample the source image while transferring high-frequency details from the guide image to the target output.  
The advantage of such data-driven methods is that learning from real image data how to optimally fuse the source and guide images can potentially give better results than a hand-crafted heuristic.
The disadvantages, as for all supervised learning, are on the one hand that one must have access to a sufficient amount of training data - in our case triplets of low-resolution source, high-resolution guide \emph{and} high-resolution target images. And on the other hand that the super-resolution algorithm is, by design, overfitted to the training data and unlikely to generalise across even mild domain shifts.
Early learning-based methods were based on the idea of dictionary learning, where an image patch is seen as a (sparse) linear combination of basis functions. For super-resolution, one jointly constructs a basis (dictionary) of corresponding source, guide and target patches, so that at test time the basis coefficients can be extracted from the source and guide images and used to reconstruct the target image \cite{Li2012,Kwon2015}.
More recently, deep convolutional networks have been used to directly learn the mapping from the two inputs to the target output, keeping the dictionary implicit in the network. The deep primal-dual network \cite{Riegler2016} employs a standard encoder-decoder architecture that takes as input the naively upsampled source image and the guide, and outputs a differential correction to the source image. The result is then further refined with non-local total variation (TV) minimisation, unrolled into a series of neural network layers. The deep joint image filter \cite{Li2016,Li2019} encodes the source image and the guide, then decodes the resulting features into the target. Most prominently, the multi-scale guided network (MSG-Net) \cite{Tak-Wai2016} extracts features at different resolutions from the guide image with an encoder branch, and uses them to guide the upsampling of the source image by concatenating them to layers of corresponding resolutions in a decoder branch that upsamples the source image. As before, the network is trained to output a differential correction of the naive upsampling. \cite{Malkin2019} targets the specific case of super-resolving semantic segmentations. The high-resolution "guide" image is passed through a standard semantic segmentation network to generate a "target" segmentation map, using a loss function that encourages the target to have the same label distribution as the low-resolution source map.
%


\section{Method}

\subsection*{Notation and Preliminaries}

\begin{figure}[t]
\begin{center}
\includegraphics[width=0.9\linewidth]{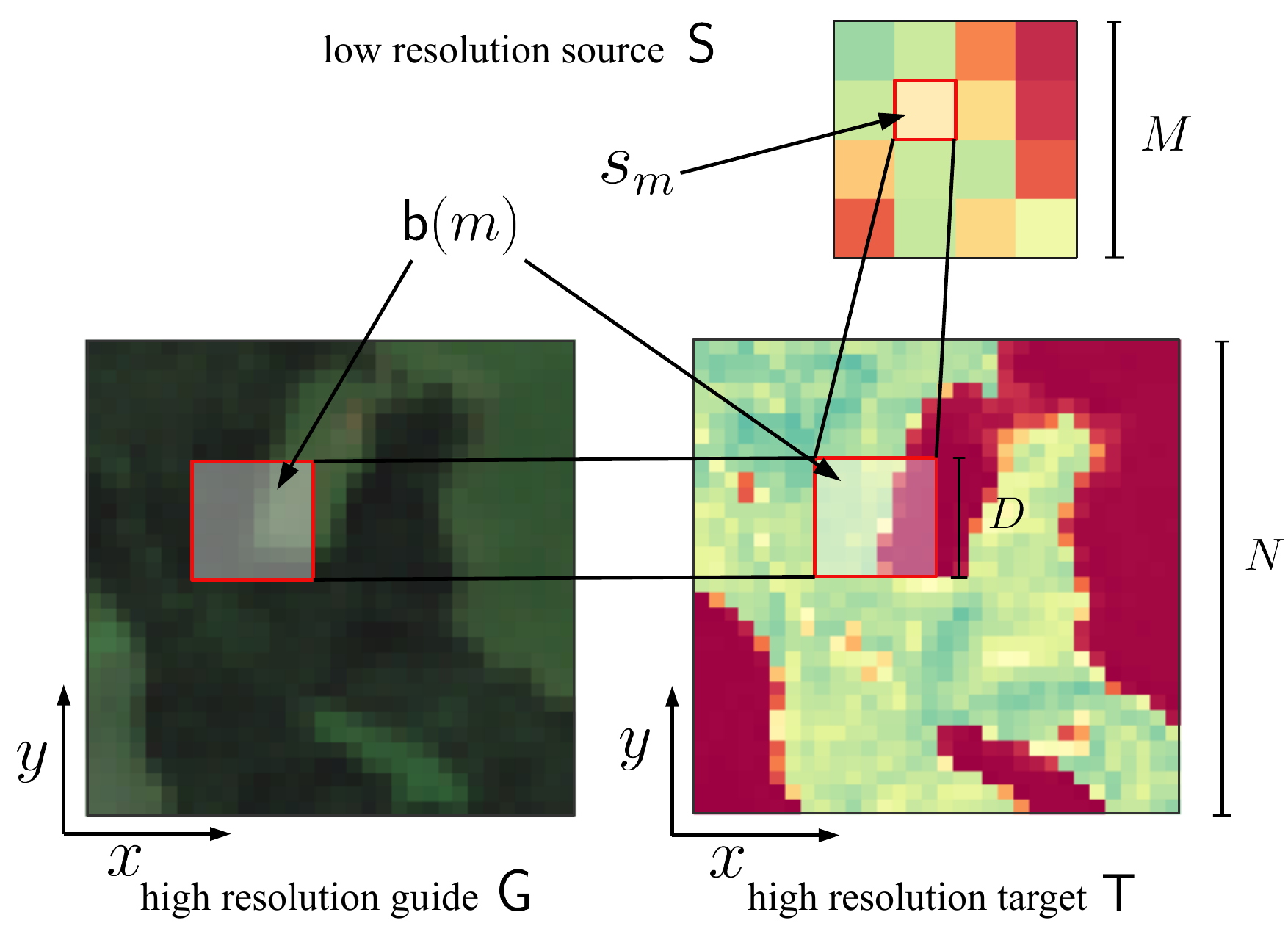}
\end{center}
   \caption{Illustration of the problem setting and notation.}
\label{fig:settings}
\end{figure}

We denote the low-resolution source map as $\mathsf{S}$, the high-resolution target map that we aim to recover as $\mathsf{T}$, and the high-resolution guide image as $\mathsf{G}$.
For simplicity, and w.l.o.g., we assume square images, with source $\mathsf{S}$ of size $M \times M$, target $\mathsf{T}$ of size $N \times N$, and guide $\mathsf{G}$ of size $N\times N\times C$, where $C$ is the number of channels. 
To simplify the notation, we use 1-dimensional pixel indices $m\in [1\hdots M^2]$ for the low resolution and $n\in [1\hdots N^2]$ for the high resolution, which can be expanded to 2-dimensional pixel coordinates $[x_m,y_m]=\mathbf{x}_m$ when needed.
The relation between $N$ and $M$ is given by the upsampling factor $D\in\mathbb{N}^+$: $N=D\cdot M$.
In other words, each source pixel $\mathsf{S}_m$ covers a block $\mathsf{b}(m)$ of $D\times D$ target pixels; see Fig.~\ref{fig:settings}.
The value of a low-resolution pixel is the average of the underlying high-resolution pixels (weighted averaging with a known point spread function is also possible, but omitted to simplify the notation):
\begin{equation}
\label{eq:consistency}
    s_m = \frac{1}{D^2}\sum_{n \in \mathsf{b}(m)} t_n = \big\langle t_n \big\rangle_{\mathsf{b}(m)}.
\end{equation}
Our goal is to obtain an estimate $\hat{\mathsf{T}}$ of the high-resolution map,
given $\mathsf{S}$ and $\mathsf{G}$.

\subsection*{Proposed Solution}

Instead of directly estimating the unknown target pixels $t_n$, we reformulate the problem as trying to find a function $f_{\boldsymbol{\theta}}:\mathbb{R}^C\rightarrow \mathbb{R}$ with parameters $\boldsymbol\theta$ that maps every guide pixel to a target pixel, $\hat{t}_n=f_{\boldsymbol{\theta}}(\mathbf{g}_n)$, such that the result is consistent with the source image according to Eq. ($\ref{eq:consistency}$).
As a loss function to measure the consistency, we empirically use the $\ell^1$-distance, leading to:
\begin{equation}
    \hat{\boldsymbol{\theta}}=\underset{\boldsymbol{\theta}}{\text{argmin}}\ \sum_{m} \big|s_m-\big\langle f_{\boldsymbol{\theta}}(\mathbf{g}_n) \big\rangle_{\mathsf{b}(m)} \big|.
    \label{eq:orig_problem}
\end{equation}
That problem is obviously ill-posed, since many different target images $\mathsf{T}$ can be constructed that have loss $0$. Moreover, even for given $\mathsf{S}$, $\mathsf{T}$ and $\mathsf{G}$ a perfect solution can always be found by choosing a sufficiently complex function%
\footnote{Except for the pathological case of two or more identical blocks in $\mathsf{G}$  with different source values.} %
$f_{\boldsymbol{\theta}}$.

\begin{figure}[t]
\begin{center}
\includegraphics[width=0.9\linewidth]{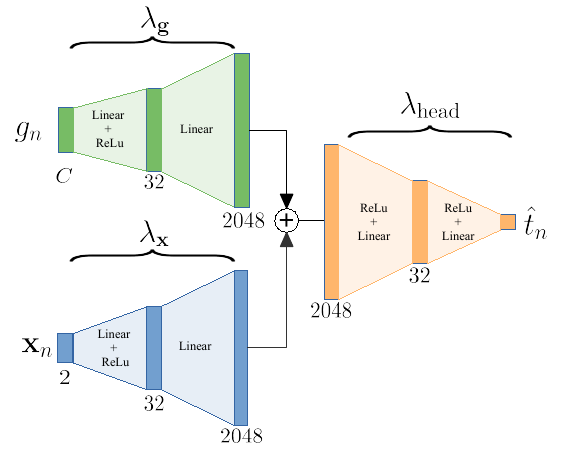}
\end{center}
   \caption{Architecture of the neural network used to model the mapping between the guide image and the high resolution map.}
\label{fig:nn}
\end{figure}

Here, we parametrise the function $f_{\boldsymbol{\theta}}$ as a multi-layer perceptron (MLP), which takes as input the $(C\times 1)$-vector of intensities at a guide pixel $\mathbf{g}_n$ and outputs the corresponding target value $\hat{t}_n$.
Restricting $f_{\boldsymbol{\theta}}$ to a function with reasonably low complexity ensures the problem is solvable. But since some input images are easier to upsample than others, always utilising the full capacity of $f_{\boldsymbol{\theta}}$ is prone to overfit in most cases.
A core insight of our method is that, instead of regularising the output $\mathsf{\hat{T}}$, one can also combat overfitting by encouraging the choice of a simpler $f_{\boldsymbol{\theta}}$ through a suitable regulariser, in our case an $\ell^2$-penalty on the network weights:
\begin{equation}
   \hat{\boldsymbol{\theta}}=\underset{\boldsymbol{\theta}}{\text{argmin}} \sum_{m} \big|s_m-\big\langle f_{\boldsymbol{\theta}}(\mathbf{g}_n) \big\rangle_{\mathsf{b}(m)} \big|+\lambda\big\|\boldsymbol{\theta}\big\| ^2\;,
    \label{eq:reg_problem}
\end{equation}
with a hyper-parameter $\lambda$ that controls the strength of the regularisation.
There is still one issue with Eq.~\eqref{eq:reg_problem}, namely that the model in this form is too restrictive: it imposes a one-to-one relationship between guide pixels $\mathbf{g}_n$ and output pixels $\hat{t}_n$. If, for instance, two pixels have the same colour in the guide, then they will be mapped to the same target depth, which is clearly not reasonable.
Our trick to inject the necessary flexibility is to additionally allow the mapping to vary across the image plane:
\begin{equation}
    \hat{\boldsymbol{\theta}}=\underset{\boldsymbol{\theta}}{\text{argmin}}\ \sum_{m} \big|s_m-\big\langle f_{\boldsymbol{\theta}}(\mathbf{g}_n,\mathbf{x}_n) \big\rangle_{\mathsf{b}(m)} \big|+\lambda\big\|\boldsymbol{\theta}\big\| ^2\;.
    \label{eq:full_problem}
\end{equation}
Note that the regulariser $\|\boldsymbol{\theta}\|^2$ enforces low complexity of the network $f_{\boldsymbol{\theta}}$ not only w.r.t.\ the guide pixel values, but also w.r.t.\ the spatial location. In practice, we found it beneficial to train separate branches for the intensities $\mathbf{g}_n$ and the coordinates $\mathbf{x}_n$, which are then merged by adding their activations, as depicted in Fig.~\ref{fig:nn}. With this architecture it is also possible to regularise each branch differently, by setting individual hyper-parameters $\lambda_\mathbf{g}, \lambda_\mathbf{x}, \lambda_\text{head}$. This is convenient when one has corresponding a priori knowledge, e.g., when super-resolving semantic segmentations one may not want the mapping to strongly vary across the image plane.

Super-resolving a given input image $\mathsf{S}$ with the help of a guide image $\mathsf{G}$ now amounts to solving the optimisation problem \eqref{eq:full_problem}. This can be done with simple stochastic gradient descent, but it may be beneficial to use more advanced optimisation schemes for this specific problem structure -- finding the best  numerical scheme is left for future work. 
Note that the optimisation over all pixels can be performed efficiently in any deep learning framework, by implementing $f_{\boldsymbol{\theta}}$ as a convolutional network $F^{1\!\times\!1}_{\boldsymbol{\theta}}$ with $(1\times 1)$ kernels on all layers. The network takes as input the complete guide image, augmented with two additional channels for the pixel indices $x_n,y_n$, and outputs the complete target image.
Once the network parameters have been fitted, the target is recovered by applying the function $f_{\boldsymbol{\theta}}$ to each pixel of the guide, which corresponds to a forward pass in the convolutional version:
\begin{equation}
\label{eq:mappingT}
\hat{t}_n=f_{\hat{\boldsymbol{\theta}}}(\mathbf{g}_n) \quad,\quad
\hat{\mathsf{T}}=F^{1\!\times\! 1}_{\hat{\boldsymbol{\theta}}}(\mathsf{G})\;.
\end{equation}

\section{Experimental Results}

In the following, we analyse the performance of the proposed pixel-to-pixel transformation method on two different datasets, and compare it to three state-of-the-art guided super-resolution methods, as well as two baselines.

\subsection*{Evaluation Settings}

In all experiments, we set the target resolution to $256^2$ pixels. We evaluate the algorithms at different 
upsampling factors, namely $\times 4$, $\times 8$, $\times 16$,  and $\times 32$, corresponding to
source resolutions of $64^2$, $32^2$, $16^2$, and $8^2$  respectively.
We test the proposed method on two different applications, super-resolving depth and super-resolving vegetation height.
For depth we use the data from the 2005 version of the Middlebury benchmark \cite{Scharstein2007, Hirschmuller2007}, from which we extract 120 high-resolution RGB images and depth maps.
For vegetation height, the test set is composed of 40 maps extracted from the Swiss national forest inventory \cite{Ginzler2015} (we use an updated version issued after the publication). As guide images we use multi-spectral images from ESA's Sentinel-2 satellite%
\footnote{Copernicus Sentinel data 2016, processed by ESA. \url{https://scihub.copernicus.eu/}}. %
The satellite sensor records 13 channels at three different resolutions, we limit ourselves to the four channels with the highest resolution of 10$\,$m per pixel, which are recorded in blue, green, red and near infra-red.
In both cases the source images are generated by downsampling the ground truth targets with the appropriate scaling factor.

As baselines we adopt on the one hand naive \emph{bicubic interpolation}, without guide image; and on the other hand the classical \emph{guided filter}~\cite{He2013}.
We further compare to two state-of-the-art methods for guided super-resolution, namely the \emph{Fast Bilateral Solver} (FBS) \cite{Barron2016} and the \emph{static-dynamic filter} (SD) ~\cite{Ham2018}. For the former we used the authors' original implementation%
\footnote{\url{https://github.com/poolio/bilateral_solver}}, %
for the latter we ported the authors' implementation%
\footnote{ \url{https://github.com/bsham/SDFilter}} %
to Python, and modified the data fidelity term of the optimisation to match the per-block consistency of Eq.~\eqref{eq:consistency}. We select the parameters of FBS and SD according to the authors' guidelines and keep them constant for all experiments. We have verified that the quantitative results are consistent with the original publications.

The last method we compare to is a recent supervised learning algorithm, \emph{MSG-Net}~\cite{Tak-Wai2016}, also in the authors' original implementation%
\footnote{\url{https://github.com/twhui/MSG-Net}}. %
%
We argue that guided super-resolution is most useful if it is not easily possible to record large amounts of data at the target resolution (e.g., large-scale vegetation height maps at 10$\,$m resolution cannot be produced at a reasonable cost). Nevertheless we included the results of this method for comparison.
Under the assumption that source and guide images are available in large quantities we follow a common procedure from the literature~\cite{Schocher2017,Lanaras2018}: under the assumption that the upsampling model is to some degree scale-invariant, one can \emph{downsample} the available $M\times M$ data by the factor $D$ to obtain synthetic training data for $\times D$ upsampling. The model thus trained for upsampling $(M/D)^2\rightarrow M^2$ is then, at test time, applied to the actual super-resolution task $M^2\rightarrow N^2$. This model will be referred to as MSG-Net-DS.
Overall, we train MSG-Net on the dataset provided by \cite{Tak-Wai2016} for the depth case and a similarly sized dataset for the vegetation height. We found that, due to the repeated downsampling, the depth data provided by \cite{Tak-Wai2016} was not enough to train the MSG-Net-DS, so we additionally used the training data of \cite{Riegler2016}, and performed some data augmentation for the vegetation height. Still, the data was not sufficient to train for factors larger than $\times 8$. For practical applications of guided super-resolution, the need for large amounts of labelled training data is a real issue, and a serious limitation.

For our method, we train the mapping $f_{\boldsymbol{\theta}}$ on batches of 32 low-resolution pixels/blocks, using the ADAM optimiser~\cite{adam} with learning rate 0.001. We centre the image values and normalise them to unit standard deviation, for both the source and the guide image. If the guide has more than one channel, we normalise them separately. Pixel coordinates $\mathbf{x}_n$ are rescaled to the interval $[-0.5,0.5]$. We train for 32'000 iterations (independent of the upsampling factor), which takes about 120 seconds for a $\times 8$ upsampling on a standard GPU (Nvidia GTX 1080 Ti). The implementation of our method is available online \footnote{\url{https://github.com/riccardodelutio/PixTransform}}.

As quantitative error metrics we use the Mean Squared Error (MSE) and the Mean Absolute Error (MAE), both in the original units of the respective datasets (pixel disparity for depth, metres for tree height).
Moreover, we also measure the Percentage of Bad Pixels (PBP) as defined in \cite{Ham2018}: 
\begin{equation}
\text{PBP}_\delta = \frac{1}{\mathrm{N}^2} \sum_n \left[|\hat{t}_n-t_n|>\delta\right]    
\end{equation}
with $\delta=1$ pixel for disparity, and $\delta=3$ metres for vegetation height. 

\subsection*{Analysis}
In this subsection we analyse the mapping learned by our method, and illustrate the influence of the regularisation.

\begin{figure}[t]
\begin{center}
\includegraphics[width=\linewidth]{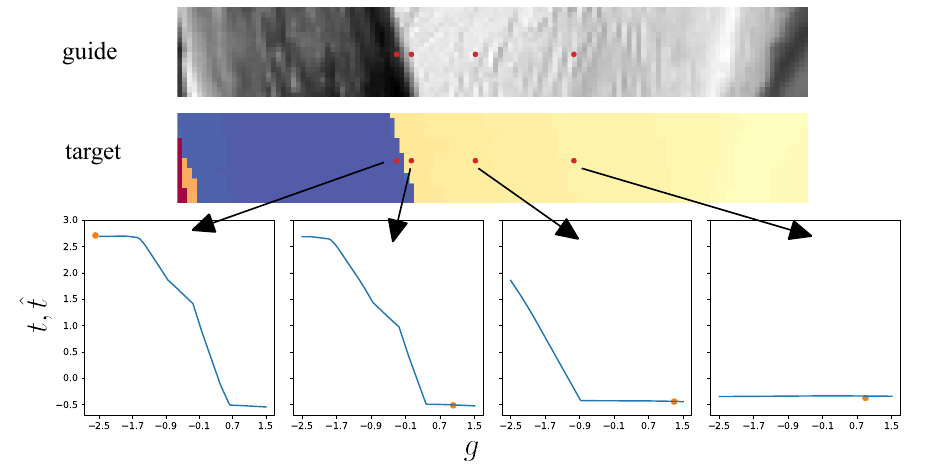}
\end{center}
   \caption{Illustration of the location-dependent mapping function $f_{\boldsymbol{\theta}}$.}
\label{fig:example}
\end{figure}

First, we visualise the mapping function $f_{\boldsymbol{\theta}}$. In Fig.~\ref{fig:example} we plot the learned dependence between intensity $g_n$ in the guide and depth $t_n$ in the target image, at different image locations $\mathbf{x}_n$.
Close to the discontinuity the function has a steep slope, as the network learns to translate the large intensity change into a large depth change, so as to be consistent with the depth change seen, at coarser resolution, in the source image.
As one moves away from the discontinuity and into the homogeneous depth region to its right, the network response flattens out, indicating that all colours of the guide shall be translated to similar depth values.
The picture nicely illustrates the mechanism behind our algorithm's ability to reproduce sharp edges: imposing smoothness on the mapping function $f_{\boldsymbol{\theta}}$ is very different from imposing smoothness on the target output. The function $f_{\boldsymbol{\theta}}$ indeed changes slowly and has similar shape at the two leftmost locations. But since that shape corresponds to a steep gradient, the depths at the two locations are very different.
Regularising the mapping function instead of the output image is a lot more robust to variations in image content.

\begin{figure}[t]
\begin{center}
\includegraphics[width=\linewidth]{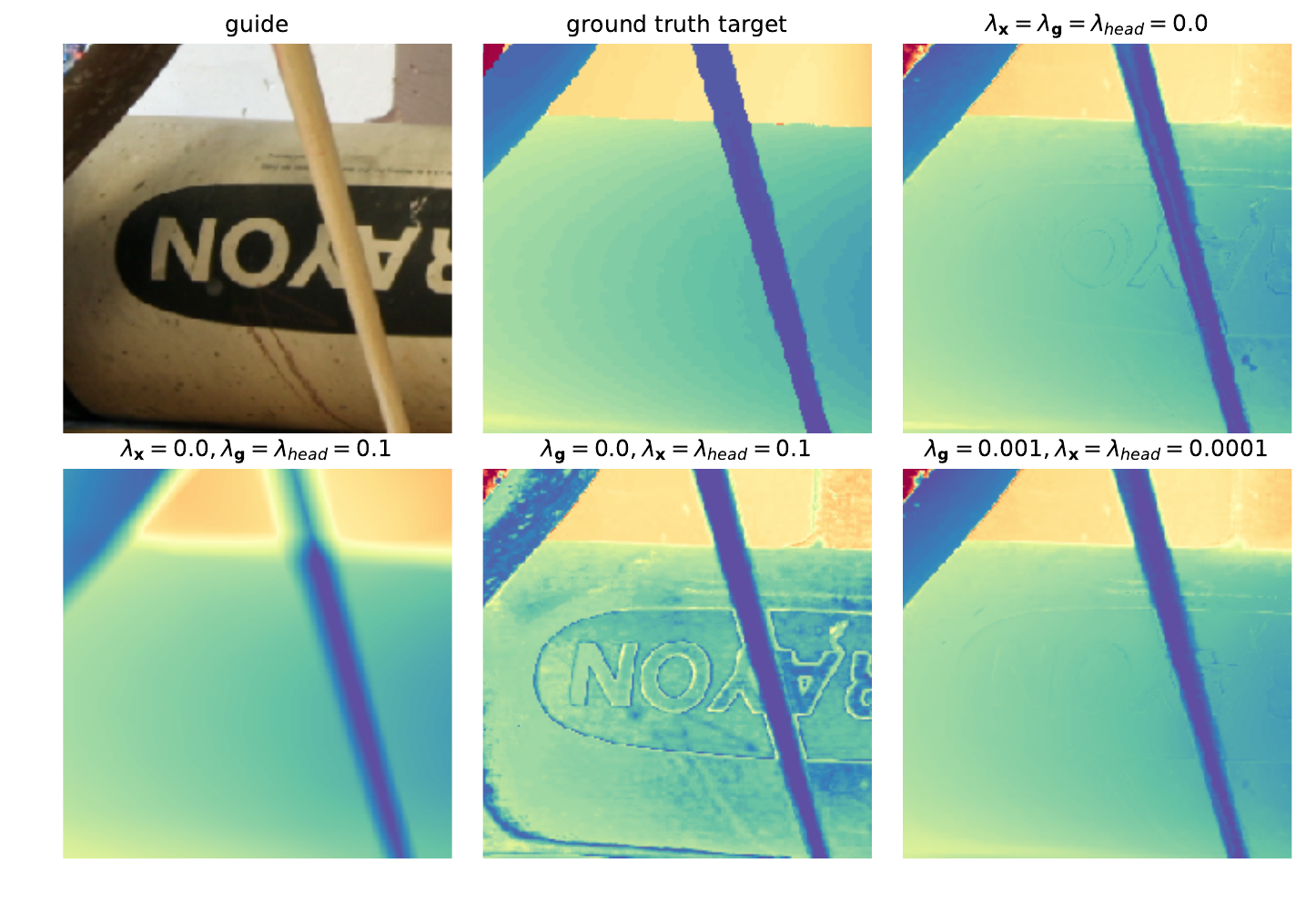}
\end{center}
   \caption{Illustration of different regularisation settings (upsampling factor $\times 8$).}
\label{fig:regs}
\end{figure}

Figure~\ref{fig:regs} depicts the effect of changing the regularisation parameters $\lambda_{\mathbf{g}}$, $\lambda_{\mathbf{x}}$ and $\lambda_{\text{head}}$.
The figure shows four cases: with no regularisation, the network $f_{\boldsymbol{\theta}}$ has more capacity than needed and overreacts to intensity contrasts in the guide. That behaviour is amplified if one excessively regularises only w.r.t.\ the location $\mathbf{x}_m$, thus forcing $f_{\boldsymbol{\theta}}$ to base its outputs mostly on the colour values $\mathbf{g}_m$. Conversely, regularising heavily only w.r.t.\ $\mathbf{g}_m$ causes the network to ignore the colours of the guide, leading to blurry outputs. In the bottom right, the regularisation weights are set to a sensible compromise: $\lambda_{\mathbf{g}} = 10^{-3}$ and $\lambda_{\mathbf{x}} = \lambda_{\text{head}} = 10^{-4}$. These are the settings used in all our experiments.

\subsection*{Depth super-resolution}

\newcounter{phase}

\renewcommand{\thephase}{\alph{phase}}

\begin{figure*}[!t]
  \centering
  \vspace{-0.5cm}
  \def\mywidth{0.152}
  \def\myheight{-0.26cm}
  \begin{tabular}{@{}c@{\hspace{1mm}}c@{\hspace{2.5mm}}c@{\hspace{1mm}}c@{\hspace{2.5mm}}c@{\hspace{1mm}}c@{\hspace{1mm}}c@{}}
    \multicolumn{2}{c}{\refstepcounter{phase} \thephase. Depth $\times 8$ \label{Depth_8}} &    
    \multicolumn{2}{c}{\refstepcounter{phase} \thephase. Depth $\times 16$ \label{Depth_16}} &        
    \multicolumn{2}{c}{\refstepcounter{phase} \thephase. Depth $\times 32$ \label{Depth_32}} \\ 
    
   \subcaptionbox*{\tiny Guide}{ \adjincludegraphics[width=\mywidth\linewidth,trim={0 0 0 0},clip]{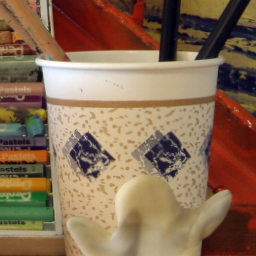}    \vspace{\myheight}} &
   \subcaptionbox*{\tiny Input}{ \adjincludegraphics[width=\mywidth\linewidth,trim={0 0 0 0},clip]{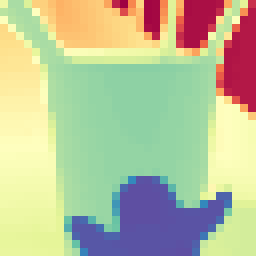}    \vspace{\myheight}} &
   \subcaptionbox*{\tiny Guide}{ \adjincludegraphics[width=\mywidth\linewidth,trim={0 0 0 0},clip]{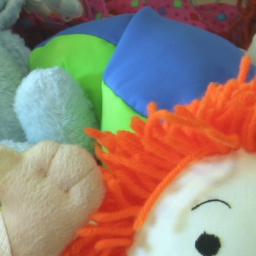}    \vspace{\myheight}} &
   \subcaptionbox*{\tiny Input}{ \adjincludegraphics[width=\mywidth\linewidth,trim={0 0 0 0},clip]{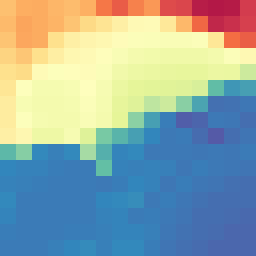}    \vspace{\myheight}} &
   \subcaptionbox*{\tiny Guide}{ \adjincludegraphics[width=\mywidth\linewidth,trim={0 0 0 0},clip]{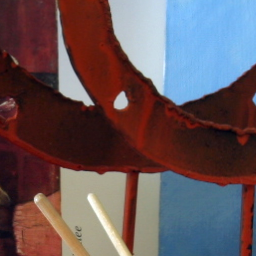}    \vspace{\myheight}} &
   \subcaptionbox*{\tiny Input}{ \adjincludegraphics[width=\mywidth\linewidth,trim={0 0 0 0},clip]{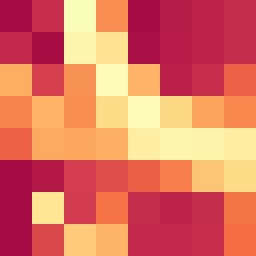}    \vspace{\myheight}} &
    \\
   \subcaptionbox*{\tiny Ours (MSE=33.3)}{ \adjincludegraphics[width=\mywidth\linewidth,trim={0 0 0 0},clip]{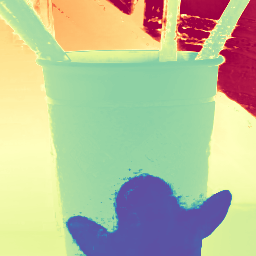}    \vspace{\myheight}} &
   \subcaptionbox*{\tiny Target}{ \adjincludegraphics[width=\mywidth\linewidth,trim={0 0 0 0},clip]{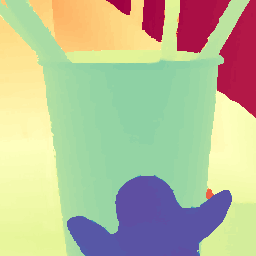}    \vspace{\myheight}} &
    \subcaptionbox*{\tiny Ours (MSE=4.6)}{ \adjincludegraphics[width=\mywidth\linewidth,trim={0 0 0 0},clip]{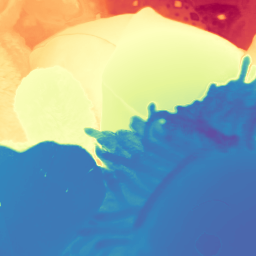}    \vspace{\myheight}} &
    \subcaptionbox*{\tiny Target}{ \adjincludegraphics[width=\mywidth\linewidth,trim={0 0 0 0},clip]{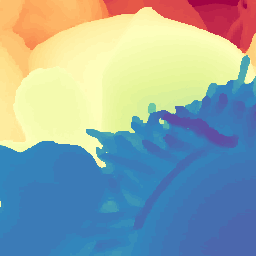}    \vspace{\myheight}} &
    \subcaptionbox*{\tiny Ours (MSE=62.6)}{ \adjincludegraphics[width=\mywidth\linewidth,trim={0 0 0 0},clip]{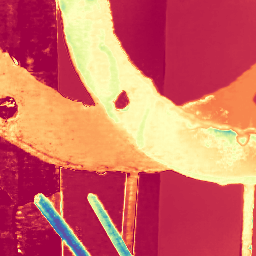}    \vspace{\myheight}} &
    \subcaptionbox*{\tiny Target}{ \adjincludegraphics[width=\mywidth\linewidth,trim={0 0 0 0},clip]{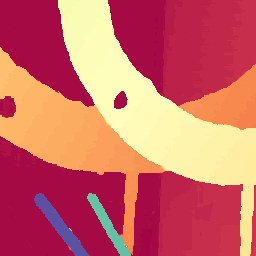}    \vspace{\myheight}} &

    \\
   \subcaptionbox*{\tiny Bicubic (MSE=40.9)}{ \adjincludegraphics[width=\mywidth\linewidth,trim={0 0 0 0},clip]{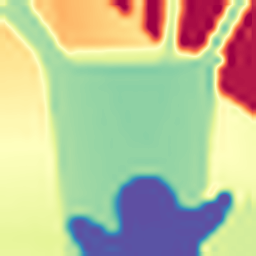}    \vspace{\myheight}} &
   \subcaptionbox*{\tiny GF (MSE=45.5)}{ \adjincludegraphics[width=\mywidth\linewidth,trim={0 0 0 0},clip]{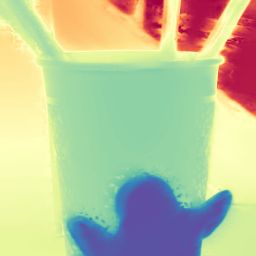}    \vspace{\myheight}} &
   \subcaptionbox*{\tiny Bicubic (MSE=16.8)}{ \adjincludegraphics[width=\mywidth\linewidth,trim={0 0 0 0},clip]{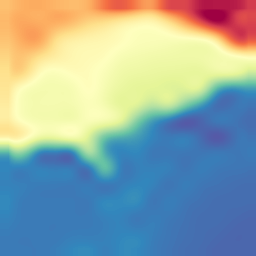}    \vspace{\myheight}} &
   \subcaptionbox*{\tiny GF (MSE=12.8)}{ \adjincludegraphics[width=\mywidth\linewidth,trim={0 0 0 0},clip]{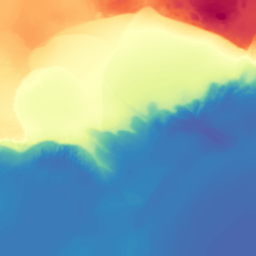}    \vspace{\myheight}} &
   \subcaptionbox*{\tiny Bicubic (MSE=246.3)}{ \adjincludegraphics[width=\mywidth\linewidth,trim={0 0 0 0},clip]{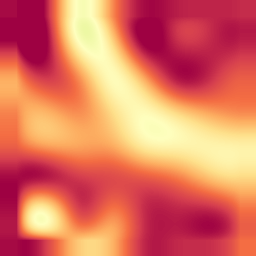}    \vspace{\myheight}} &
   \subcaptionbox*{\tiny GF (MSE=234.3)}{ \adjincludegraphics[width=\mywidth\linewidth,trim={0 0 0 0},clip]{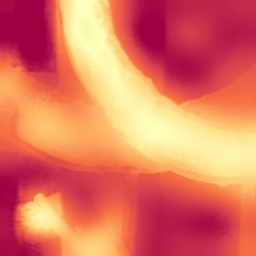}    \vspace{\myheight}} &

    \\
   \subcaptionbox*{\tiny FBS (MSE=73.4)}{ \adjincludegraphics[width=\mywidth\linewidth,trim={0 0 0 0},clip]{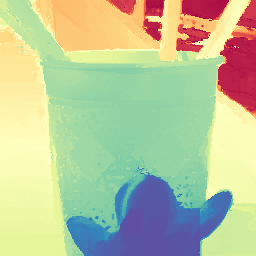}    \vspace{\myheight}} &
   \subcaptionbox*{\tiny SD Filter (MSE=46.3)}{ \adjincludegraphics[width=\mywidth\linewidth,trim={0 0 0 0},clip]{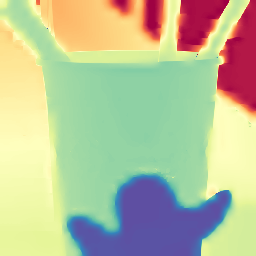}    \vspace{\myheight}} &
   \subcaptionbox*{\tiny FBS (MSE=12.3)}{ \adjincludegraphics[width=\mywidth\linewidth,trim={0 0 0 0},clip]{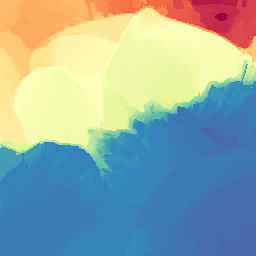}    \vspace{\myheight}} &
   \subcaptionbox*{\tiny SD Filter (MSE=18.4)}{ \adjincludegraphics[width=\mywidth\linewidth,trim={0 0 0 0},clip]{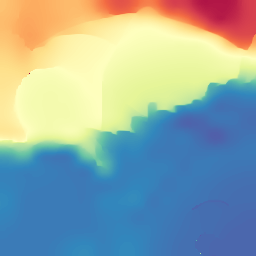}    \vspace{\myheight}} &
   \subcaptionbox*{\tiny FBS (MSE=205.8)}{ \adjincludegraphics[width=\mywidth\linewidth,trim={0 0 0 0},clip]{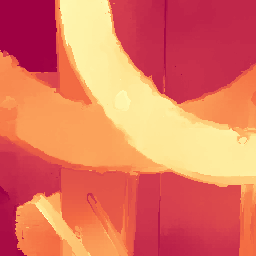}    \vspace{\myheight}} &
   \subcaptionbox*{\tiny SD Filter (MSE=20795.8)}{ \adjincludegraphics[width=\mywidth\linewidth,trim={0 0 0 0},clip]{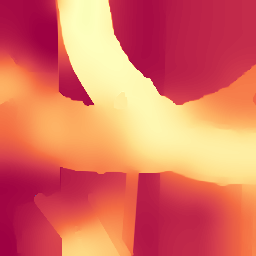}    \vspace{\myheight}} &

    \\
   \subcaptionbox*{\tiny $\text{MSG-Net-DS}$ (MSE=23.0)}{ \adjincludegraphics[width=\mywidth\linewidth,trim={0 0 0 0},clip]{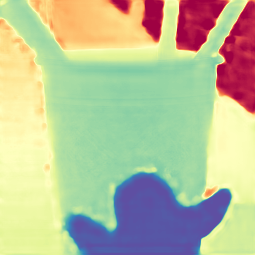}    \vspace{\myheight}} &
   \subcaptionbox*{\tiny $\text{MSG-Net}$ (MSE=15.6)}{ \adjincludegraphics[width=\mywidth\linewidth,trim={0 0 0 0},clip]{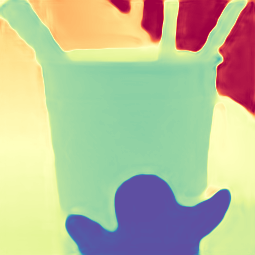}    \vspace{\myheight}} &
    \multicolumn{2}{c}{  \subcaptionbox*{\tiny $\text{MSG-Net}$ (MSE=11.4)}{ \adjincludegraphics[width=\mywidth\linewidth,trim={0 0 0 0},clip]{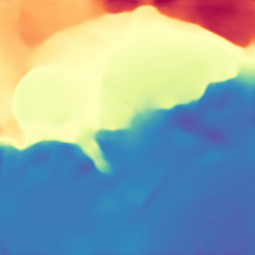}    \vspace{\myheight}}} &
    \multicolumn{2}{c}{  \subcaptionbox*{\tiny $\text{MSG-Net}$ (MSE=213.1)}{ \adjincludegraphics[width=\mywidth\linewidth,trim={0 0 0 0},clip]{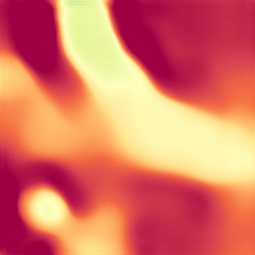}    \vspace{\myheight}}} &

    \\  

\end{tabular}
\caption{Qualitative results of different methods for depth guided super-resolution.}
\label{fig:cmp_grid}
\end{figure*}

\newcounter{phas}

\renewcommand{\thephas}{\alph{phas}}

\begin{figure*}[!t]
  \centering
  \vspace{-0.5cm}
  \def\mywidth{0.152}
  \def\myheight{-0.26cm}
  \begin{tabular}{@{}c@{\hspace{1mm}}c@{\hspace{2.5mm}}c@{\hspace{1mm}}c@{\hspace{2.5mm}}c@{\hspace{1mm}}c@{\hspace{1mm}}c@{}}
    \multicolumn{2}{c}{\refstepcounter{phas} \thephas. Vegetation $\times 8$ \label{Vegetation_8}} &
    \multicolumn{2}{c}{\refstepcounter{phas} \thephas. Vegetation $\times 16$ \label{Vegetation_16}} &     
    \multicolumn{2}{c}{\refstepcounter{phas} \thephas. Vegetation $\times 32$ \label{Vegetation_32}}  
    \\
    
   \subcaptionbox*{\tiny Guide}{ \adjincludegraphics[width=\mywidth\linewidth,trim={0 0 0 0},clip]{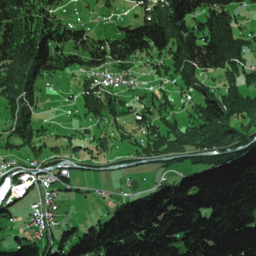}    \vspace{\myheight}} &
   \subcaptionbox*{\tiny Input}{ \adjincludegraphics[width=\mywidth\linewidth,trim={0 0 0 0},clip]{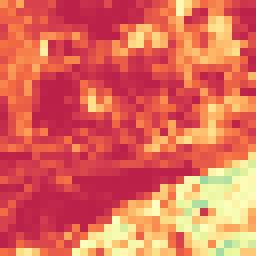}    \vspace{\myheight}} &
   \subcaptionbox*{\tiny Guide}{ \adjincludegraphics[width=\mywidth\linewidth,trim={0 0 0 0},clip]{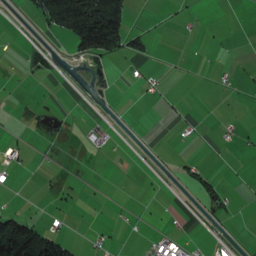}    \vspace{\myheight}} &
   \subcaptionbox*{\tiny Input}{ \adjincludegraphics[width=\mywidth\linewidth,trim={0 0 0 0},clip]{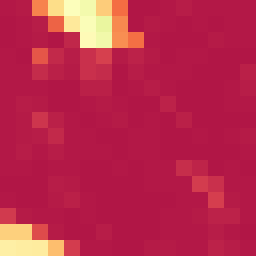}    \vspace{\myheight}} &
   \subcaptionbox*{\tiny Guide}{ \adjincludegraphics[width=\mywidth\linewidth,trim={0 0 0 0},clip]{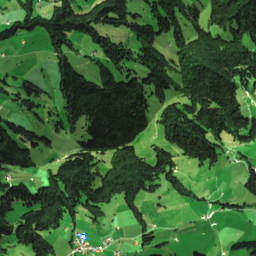}    \vspace{\myheight}} &
   \subcaptionbox*{\tiny Input}{ \adjincludegraphics[width=\mywidth\linewidth,trim={0 0 0 0},clip]{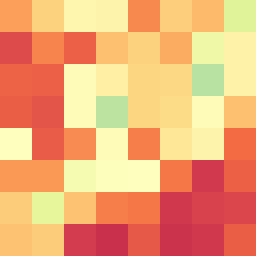}    \vspace{\myheight}} &
   
    \\
   \subcaptionbox*{\tiny Ours (MSE=15.5)}{ \adjincludegraphics[width=\mywidth\linewidth,trim={0 0 0 0},clip]{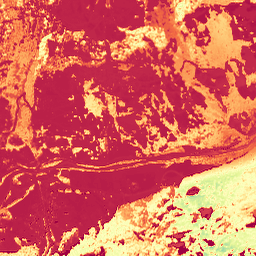}    \vspace{\myheight}} &
   \subcaptionbox*{\tiny Target}{ \adjincludegraphics[width=\mywidth\linewidth,trim={0 0 0 0},clip]{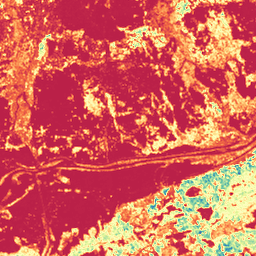}    \vspace{\myheight}} &
   \subcaptionbox*{\tiny Ours (MSE=3.0)}{\adjincludegraphics[width=\mywidth\linewidth,trim={0 0 0 0},clip]{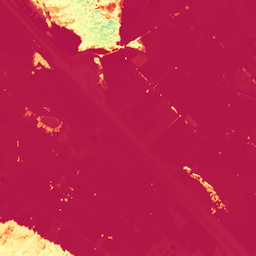}    \vspace{\myheight}} &
   \subcaptionbox*{\tiny Target}{ \adjincludegraphics[width=\mywidth\linewidth,trim={0 0 0 0},clip]{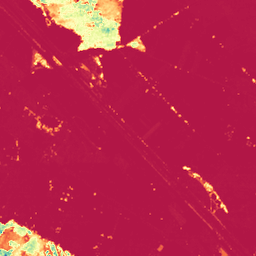}    \vspace{\myheight}} &
   \subcaptionbox*{\tiny Ours (MSE=33.8)}{ \adjincludegraphics[width=\mywidth\linewidth,trim={0 0 0 0},clip]{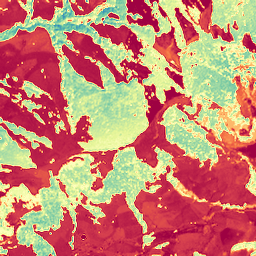}    \vspace{\myheight}} &
   \subcaptionbox*{\tiny Target}{ \adjincludegraphics[width=\mywidth\linewidth,trim={0 0 0 0},clip]{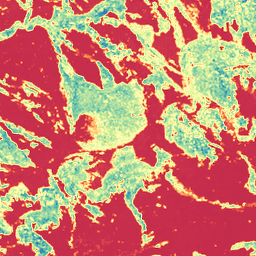}    \vspace{\myheight}} &

    \\
   \subcaptionbox*{\tiny Bicubic (MSE=19.8)}{ \adjincludegraphics[width=\mywidth\linewidth,trim={0 0 0 0},clip]{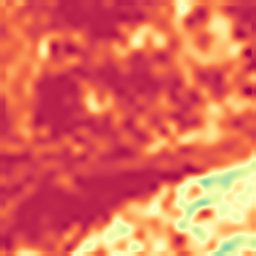}    \vspace{\myheight}} &
   \subcaptionbox*{\tiny GF (MSE=20.4)}{ \adjincludegraphics[width=\mywidth\linewidth,trim={0 0 0 0},clip]{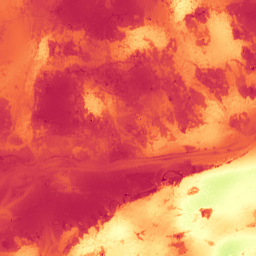}    \vspace{\myheight}} &
   \subcaptionbox*{\tiny Bicubic (MSE=7.5)}{ \adjincludegraphics[width=\mywidth\linewidth,trim={0 0 0 0},clip]{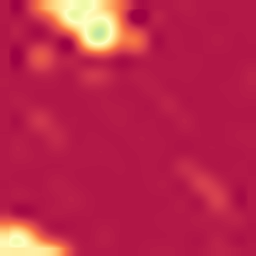}    \vspace{\myheight}} &
   \subcaptionbox*{\tiny GF (MSE=6.3)}{ \adjincludegraphics[width=\mywidth\linewidth,trim={0 0 0 0},clip]{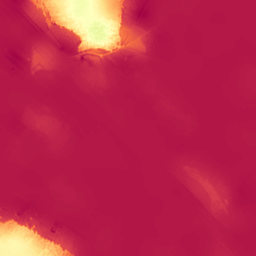}    \vspace{\myheight}} &
   \subcaptionbox*{\tiny Bicubic (MSE=89.8)}{ \adjincludegraphics[width=\mywidth\linewidth,trim={0 0 0 0},clip]{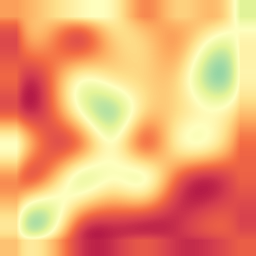}    \vspace{\myheight}} &
   \subcaptionbox*{\tiny GF (MSE=87.8)}{ \adjincludegraphics[width=\mywidth\linewidth,trim={0 0 0 0},clip]{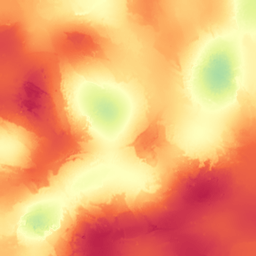}    \vspace{\myheight}} &

    \\
   \subcaptionbox*{\tiny FBS (MSE=25.6)}{ \adjincludegraphics[width=\mywidth\linewidth,trim={0 0 0 0},clip]{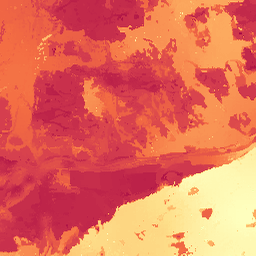}    \vspace{\myheight}} &
   \subcaptionbox*{\tiny SD Filter (MSE=21.5)}{ \adjincludegraphics[width=\mywidth\linewidth,trim={0 0 0 0},clip]{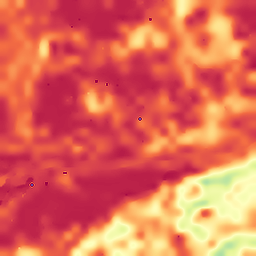}    \vspace{\myheight}} &
   \subcaptionbox*{\tiny FBS (MSE=7.8)}{ \adjincludegraphics[width=\mywidth\linewidth,trim={0 0 0 0},clip]{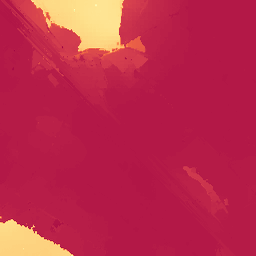}    \vspace{\myheight}} &
   \subcaptionbox*{\tiny SD Filter (MSE=16.6)}{ \adjincludegraphics[width=\mywidth\linewidth,trim={0 0 0 0},clip]{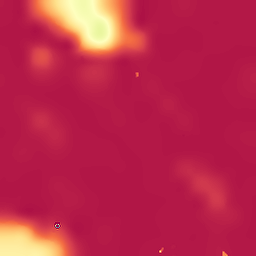}    \vspace{\myheight}} &
   \subcaptionbox*{\tiny FBS (MSE=114.1)}{ \adjincludegraphics[width=\mywidth\linewidth,trim={0 0 0 0},clip]{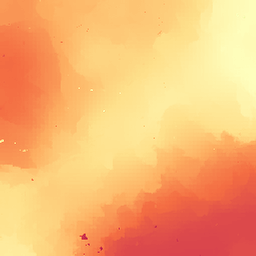}    \vspace{\myheight}} &
   \subcaptionbox*{\tiny SD Filter (MSE=226.3)}{ \adjincludegraphics[width=\mywidth\linewidth,trim={0 0 0 0},clip]{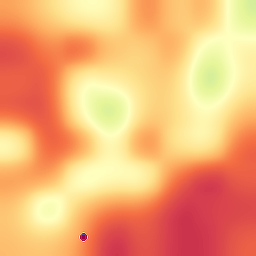}    \vspace{\myheight}} &

    \\
   \subcaptionbox*{\tiny $\text{MSG-Net-DS}$ (MSE=20.0)}{ \adjincludegraphics[width=\mywidth\linewidth,trim={0 0 0 0},clip]{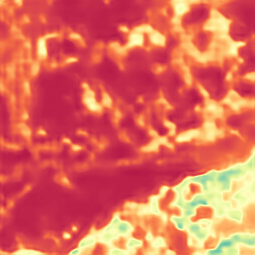}    \vspace{\myheight}} &
   \subcaptionbox*{\tiny $\text{MSG-Net}$ (MSE=18.3)}{ \adjincludegraphics[width=\mywidth\linewidth,trim={0 0 0 0},clip]{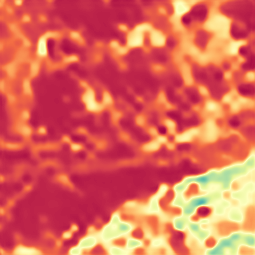}    \vspace{\myheight}} &
   \multicolumn{2}{c}{  \subcaptionbox*{\tiny $\text{MSG-Net}$ (MSE=7.3)}{ \adjincludegraphics[width=\mywidth\linewidth,trim={0 0 0 0},clip]{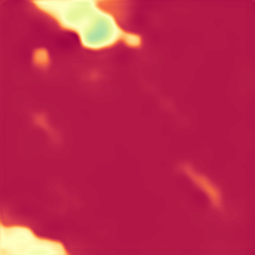}    \vspace{\myheight}}} &
   \multicolumn{2}{c}{  \subcaptionbox*{\tiny $\text{MSG-Net}$ (MSE=103.9)}{ \adjincludegraphics[width=\mywidth\linewidth,trim={0 0 0 0},clip]{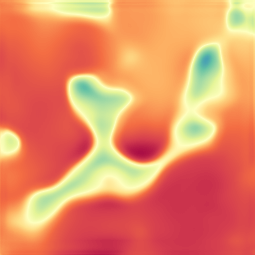}    \vspace{\myheight}}} &

    \\  

\end{tabular}
\caption{Qualitative results of different methods for vegetation height guided super-resolution.}

\label{fig:cmp_grid_vhm}
\end{figure*}

As commonly done, we run the super-resolution in disparity (inverse depth) space. In Table~\ref{tab:cmp_depth} we show the means and standard deviations of the three error metrics MSE, MAE and PBP over the images in the depth dataset, for upsampling factors of $\times 4$, $\times 8$, $\times 16$ and $\times 32$. 

Overall the MSG-Net performs well, especially for small upsampling factors. Nevertheless, as the main assumption about data availability is drastically different than ours, thus we will focus our analysis only on the methods that don't use high-res ground truth target data for training.

For a $\times 4$ upsampling factor all methods achieve similar performance. MSG-Net-DS stands out for having very low MSE, probably since it was optimised on a huge training set to minimise that error. The SD filter has a slight edge in terms of robustness and reaches the lowest MAE and PBP. It is worth pointing out that even naive bicubic upsampling is competitive, i.e., upsampling by a moderate $\times 4$ is quite an easy problem, for which the guide image has only limited effect.

For larger upsampling factors our method outperforms all others w.r.t.\ all three metrics. We could not run MSG-Net-DS for factors above $\times 8$, because not enough training data was left after downsampling the low-resolution source images. 

Fig.~\ref{fig:cmp_grid}\ref{Depth_8} shows a depth upsampling result for upsampling factor $\times 8$. Although our method on average achieves the best results for this task -- see Tab.~\ref{tab:cmp_depth} -- we deliberately show an image where MSG-Net-DS has lower MSE. Nevertheless, our output is visibly sharper and better preserves discontinuities. The top right corner of the image shows a particularly difficult situation where the contrast is high, and nearby pixels have similar colours, but different depths. In this situation several methods, including ours, exhibit a tendency to rely too much on the guide image and hallucinate spurious depth patterns. In such cases, an additional regularisation of the output, e.g., with a total variation prior, could potentially be helpful.

Fig.~\ref{fig:cmp_grid}\ref{Depth_32} shows the results for depth upsampling by a factor $\times 32$. As can be seen, our method greatly outperforms the competitors. Not only it achieves a much lower MSE, but also the resulting image is a lot sharper and exhibits fewer artefacts.
Notice in particular the two thin sticks at the bottom, where only our method reaches a reasonable reconstruction quality.
Another impressive feature is the reconstruction of the hole in the middle of the image. While it is not that surprising that the boundary can be transferred from the guide; it is remarkable that from seeing the red colour of the foreground, the white colour in the background outside the object, and the area-weighted depth average of the two in the source, the network is able to extract enough information to choose the correct depth in the hole. 

\begin{figure}[h]
\begin{center}
\includegraphics[width=\linewidth]{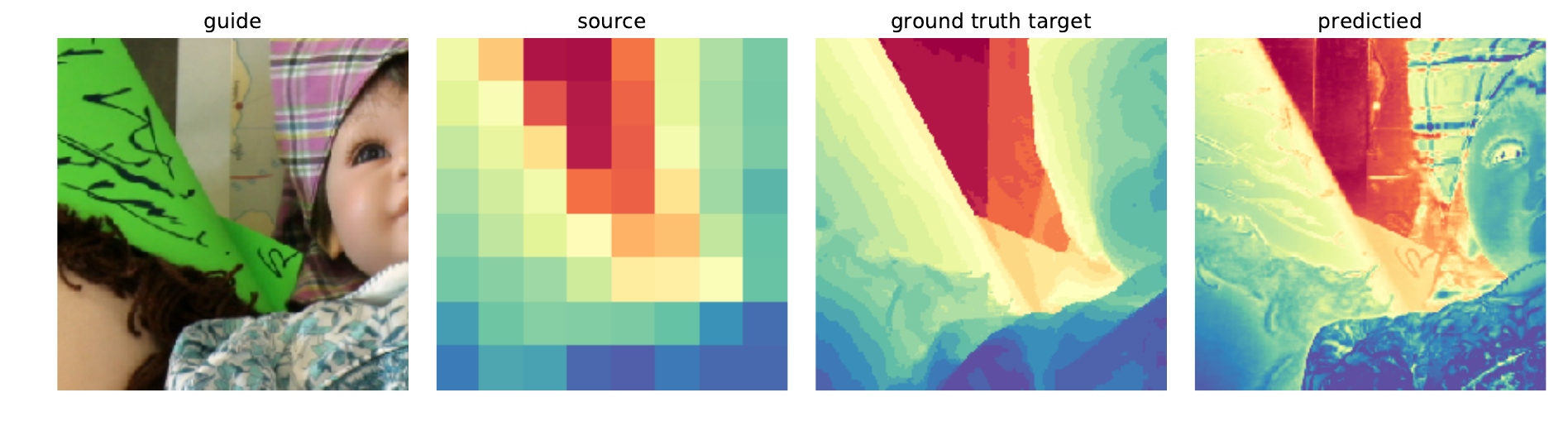}
\vspace{-1em}
   \caption{An example of $\times 32$ super-resolution where our method fails. The predicted target is corrupted with lots of high-frequency details from the highly textured guide.}
\label{fig:bad}
\end{center}
\end{figure}

Super-resolution by a factor as high as $\times 32$ is evidently pushing things to the limit of what is possible, and satisfactory results are not reached for all images. Figure~\ref{fig:bad} shows a failure case. The guide image has a lot of texture details, and nearby pixels with the same colour but different depths. The target is still consistent with the source and contains the true depth boundaries, but our method also transfers a lot of spurious texture details where there should not be depth discontinuities.
It may be possible to mitigate the problem -- but probably not completely solve -- by stronger regularisation, perhaps making the regulariser $\lambda_{\mathbf{g}}$ dependent on the upsampling factor.

\begin{table*}[t!]
\centering
\resizebox{\textwidth}{!}{%
\begin{tabular}{ccccccccc}
                             &     & Bicubic          & GF \cite{He2013}    & FBS    \cite{Barron2016}         & SD filter \cite{Ham2018}                & $\text{MSG-Net-DS}$ \cite{Tak-Wai2016} & $\text{MSG-Net}^\dagger$ \cite{Tak-Wai2016}   & Ours                                      \\ \hline \hline
\multirow{3}{*}{$\times 4$}  & MSE & 6.5 (11.5)     & 7.3 (13.0)    & 6.6 (10.9)    & 5.5 (9.9)              & \textbf{1.9 (3.0)}        & {\cellcolor{Gray}} 1.6 (3.0)            & 5.0 (8.6)                              \\ \hhline{~--------}
                             & MAE & 0.6 (0.5)    & 0.8 (0.6)   & 0.8 (0.5)   & \textbf{0.4 (0.4)}   & \textbf{0.4 (0.2)}    & {\cellcolor{Gray}} 0.3 (0.2)                  & 0.5 (0.3)                             \\ \hhline{~--------}
                             & $\text{PBP}_{\delta=1}$ & 7.5  (5.8)     & 12.3 (8.4)    & 14.3 (9.4)    &{\cellcolor{Gray}} \textbf{4.5 (3.8)}     & 6.0 (4.9)    & 5.3 (4.6)                            & 6.9 (5.1)                           \\ \hline\hline
\multirow{3}{*}{$\times 8$}  & MSE & 12.2 (21.9)    & 10.2 (18.5)   & 11.9 (18.5)   & 15.1 (27.4)            & 8.3 (11.2)       &\cellcolor{Gray} 4.1 (7.8)                    & \textbf{5.6 (9.7)}        \\ \hhline{~--------}
                             & MAE & 1.0 (0.9)    & 1.0 (0.9)   & 1.3 (0.9)   & 0.7 (0.7)            & 1.4 (0.5)          & \cellcolor{Gray} 0.6 (0.4)                   &\cellcolor{Gray} \textbf{0.6 (0.4)}           \\ \hhline{~--------}
                             & $\text{PBP}_{\delta=1}$ & 14.6 (10.0)    & 16.3 (10.8)   & 29.9 (16.6)   & 9.1 (7.1)              & 43.7 (8.5)      & 11.3 (8.5)                       & \cellcolor{Gray}\textbf{8.8 (6.8)}             \\ \hline\hline
\multirow{3}{*}{$\times 16$} & MSE & 26.5 (48.7)    & 21.6 (40.9)   & 19.3 (34.9)   & 115.5 (369.7)          & -              & 12.4 (26.7)                        & \cellcolor{Gray}\textbf{8.4 (14.9)}            \\ \hhline{~--------}
                             & MAE & 1.9 (1.8)    & 1.7 (1.6)   & 1.8 (1.5)   & 1.3 (1.5)            & -                & 1.2 (1.0)                     & \cellcolor{Gray}\textbf{0.9 (0.7)}           \\ \hhline{~--------}
                             & $\text{PBP}_{\delta=1}$ & 27.3 (15.8)    & 26.8 (15.4)   & 38.8 (19.3)   & 18.7 (12.5)            & -                  & 24.3 (13.6)                   & \cellcolor{Gray}\textbf{15.5 (10.9)}           \\ \hline\hline
\multirow{3}{*}{$\times 32$} & MSE & 54.1 (95.2)    & 49.7 (88.3)   & 40.2 (72.3)   & 1343.3 (3374.5)          & -                  & 42.5 (79.8)             &\cellcolor{Gray} \textbf{26.0 (42.9)}            \\ \hhline{~--------}
                             & MAE &  3.3 (2.9)    & 3.2 (2.8)   & 3.0 (2.5)   & 2.7 (2.6)           & -                         & 2.8 (2.3)              &\cellcolor{Gray} \textbf{2.0 (1.7)}           \\ \hhline{~--------}
                             & $\text{PBP}_{\delta=1}$ & 44.9 (21.6)    & 45.0 (21.7)   & 50.6 (22.5)   & 37.2 (19.4)           & -                    & 46.2 (20.5)                   &\cellcolor{Gray} \textbf{36.3 (20.6)}           \\ \hline\hline
\end{tabular}%
}
\vspace{-0.5em}
\caption{Performance comparison with the state-of-the-art algorithms on the depth map dataset for different values of upsampling factors. The tables shows the means and (standard deviations) over all images of the MSE (in pixel$^2$), MAE (in pixels), and PBP (in \%). $\dagger$ trained on the high-res ground truth target.\colorbox{Gray}{Best overall}, \textbf{Best without high-res ground truth targets}.}
\label{tab:cmp_depth}
\end{table*}

\begin{table*}[t!]
\centering
\resizebox{\textwidth}{!}{%
\begin{tabular}{ccccccccc}
                             &     & Bicubic          & GF \cite{He2013}    & FBS    \cite{Barron2016}         & SD filter \cite{Ham2018}                & $\text{MSG-Net-DS}$ \cite{Tak-Wai2016} & $\text{MSG-Net}^\dagger$ \cite{Tak-Wai2016}   & Ours                                      \\ \hline \hline

\multirow{3}{*}{$\times 8$}  & MSE &18.1 (13.3)           & 19.0 (14.1)   & 28.2 (24.8)                   & 20.7 (15.8)          &      17.9 (13.3)      &\cellcolor{Gray} {16.3 (12.1)}         & \textbf{17.6 (15.1)}       \\ \hhline{~--------}
                             & MAE & 2.4 (1.5)          & 2.5 (1.7)         & 3.1 (2.2)                   & 2.4 (1.6)          &      2.3 (1.5)         &\cellcolor{Gray} {2.1 (1.4)}      &\cellcolor{Gray} \textbf{2.1 (1.5)} \\ \hhline{~--------}
                             & $\text{PBP}_{\delta=3}$ & 26.2 (19.2)        & 28.2 (21.4)             & 32.3 (25.0)                 & 26.8 (19.8)        &      26.1 (19.3)     &\cellcolor{Gray} {22.7 (16.9)}         & \textbf{23.5 (18.2)} \\ \hline \hline

\multirow{3}{*}{$\times 16$} & MSE & 29.1 (22.5)          & 27.7 (21.1) & 33.7 (27.8) & 45.1 (45.4)          & -                & 29.2 (22.9)                &\cellcolor{Gray} \textbf{19.7 (17.2)} \\ \hhline{~--------}
                             & MAE & 3.1 (2.1)          & 3.1 (2.1) & 3.5 (2.5) & 3.8 (2.2)          & -                     & 2.9 (2.0)           &\cellcolor{Gray} \textbf{2.3 (1.7)} \\ \hhline{~--------}
                             & $\text{PBP}_{\delta=3}$ & 33.0 (24.4)          & 33.7 (25.5) & 36.9 (28.0) & 34.2 (25.6)          & -              & 28.3 (20.9)                  &\cellcolor{Gray} \textbf{24.2 (18.9)} \\ \hline \hline
                             
\multirow{3}{*}{$\times 32$}  & MSE &  41.5 (33.6)         & 40.2 (32.6)  & 42.3 (34.4) & 160.0 (228.3)          &  -                 & 44.3 (37.1)             &\cellcolor{Gray} \textbf{21.2 (17.5)  }        \\ \hhline{~--------}
                             & MAE &  4.0 (2.8)        & 3.9 (2.8) & 4.1 (2.9) & 4.2 (3.0)        &  -                 & 3.8 (2.7)             & \cellcolor{Gray} \textbf{2.6 (1.8) }      \\ \hhline{~--------}
                             & $\text{PBP}_{\delta=3}$ & 39.1 (29.4)      & 39.3 (29.8) & 42.0 (31.8) &  40.9 (30.7)          &  -           & 34.0 (26.3)                   & \cellcolor{Gray}\textbf{29.2 (22.4)}        \\ \hline \hline
\end{tabular}%
}
\vspace{-0.5em}
\caption{Performance comparison with the state-of-the-art algorithms on the vegetation height map dataset for different values of upsampling factors. The tables shows the means and (standard deviations) over all images of the MSE (in m$^2$), MAE (in m), and PBP (in \%). $\dagger$ trained on the high-res ground truth target. \colorbox{Gray}{Best overall}, \textbf{Best without high-res ground truth targets}.}
\label{tab:cmp_vegetation}
\end{table*}

\subsection*{Super-resolution of vegetation height}

Table~\ref{tab:cmp_vegetation} again shows the means and standard deviations of the three error metrics over the images of the vegetation height dataset, for upsampling factors $\times 8$, $\times 16$, and $\times 32$.
On this dataset most methods, including the bicubic upsampling, still have comparable performance at upsampling factor $\times 8$, likely because  vegetation height maps are in general smoother than depth maps. Visually, our method is again clearly sharper and recovers more high-frequency details than its competitors, see Fig.~\ref{fig:cmp_grid_vhm}\ref{Vegetation_8}-\ref{Vegetation_32}.
As for the depth case, our method outperforms others by a considerable margin at higher upsampling factors, in all three metrics.

Fig.~\ref{fig:cmp_grid_vhm}\ref{Vegetation_8} shows the results for vegetation upsampling by a factor $\times 8$. While the MSE values are not that different, there is nevertheless a pronounced qualitative difference between our method and all others. The one that comes closest is MSG-Net-DS, but even after having seen thousands of low-res / high-res pairs during training, the network is not able to fully recover the high-frequency details and misses a lot of the fine structures. FBS produces fairly sharp discontinuities, but has a bias towards piece-wise constant outputs, such that many of the fine details are also lost.
In a sense, all methods except for ours fail, in that they perform similar to naive bicubic interpolation without a guide image, or even worse.

Fig.~\ref{fig:cmp_grid_vhm}\ref{Vegetation_32} shows an example for the extreme case of $\times 32$ upsampling. The example illustrates that methods which start by blowing up the low-resolution source image cannot bridge such large resolution differences and essentially produce a smoothed version of the input.
On the contrary, our method, which relies more strongly on the guide image, shines in this difficult scenario. In the pixel-to-pixel transformation from the image domain to  vegetation height, no spatial detail is lost. While it appears that even the average values over large blocks of $32\times 32$ pixels provide enough information to constrain the super-resolution in the target domain.
Obviously, it depends also on the nature of the images 
whether such extreme super-resolution is feasible. In the case of the remote sensing images, the function $f_{\boldsymbol{\theta}}$ is mostly driven by the colours $\mathbf{g}_n$ of the guide, with only little spatial variation.
Still, while it is less surprising that the height 0$\,$m is correctly recovered outside the forest, which largely corresponds to a semantic segmentation of the guide; it is pleasing that within the forest regions a large portion of the height variability is correctly reconstructed too (yellow to green tones in Fig.~\ref{fig:cmp_grid_vhm}\ref{Vegetation_32}).







\section{Conclusions}
We have proposed a novel, unsupervised method for guided super-resolution.  The key idea is to view the problem as a pixel-wise transformation of the high-res guide image to the domain of the low-res source image. By choosing a multi-layer perceptron as mapping function, inference in our model is the same as fitting a CNN with only $(1\times 1)$ kernels to the guide, where the loss function is the compatibility between the downsampled output and the source image.
The advantage of our model is that, by construction, it avoids all unnecessary blurring. On the one hand, it does not involve any upsampling of the source image by interpolation. On the other hand, the reconstruction of the super-resolved target image is regularised at the level of the mapping function, in the spirit of CNNs, by fitting the same kernels to tens of thousands of pixels, and by penalising large weights (weight decay).
Consequently, our method is able to recover very fine structures and extremely sharp edges even at high upsampling factors, setting a new state of the art.

In future work, we hope to extend the approach to handle not only super-resolution of coarse source images, but also inpainting of sparse source images, so as to recover for instance vegetation height from sparse field samples.

{\small
\newpage
\bibliographystyle{ieee}
\bibliography{mybib}
}

\end{document}